\documentclass{article}

\usepackage[utf8]{inputenc}
\usepackage{amssymb,amsmath}
\usepackage{pifont}
\usepackage{graphicx,verbatimbox}
\usepackage{hyperref} 
\usepackage{tcolorbox}

\usepackage{xspace}
\usepackage{booktabs}
\usepackage{multirow}
\usepackage{rotating}
\usepackage{color}
\usepackage{palatino}

\newcommand{\stdminus}[1]{ \scalebox{0.65}{$\pm #1$}}

\newcommand{\cmark}{\ding{51}\xspace}%
\newcommand{\cmarkg}{\textcolor{lightgray}{\ding{51}}\xspace}%
\newcommand{\xmark}{\ding{55}\xspace}%
\newcommand{\xmarkg}{\textcolor{lightgray}{\ding{55}}\xspace}%

\newcommand{\adamwg}{\textcolor{lightgray}{adamw}}

\def \ours          {DeiT\xspace}
\def \ourstiny      {DeiT-Ti\xspace}
\def \ourssmall     {DeiT-S\xspace}
\def \oursbase      {DeiT-B\xspace}

\def \ourstinydis      {DeiT-Ti\alambic\xspace}
\def \ourssmalldis     {DeiT-S\alambic\xspace}
\def \oursbasedis   {DeiT-B\alambic\xspace}
\def \oursbaseup    {DeiT-B$\uparrow$}

\def \oursbasedisup {DeiT-B\alambic$\uparrow$}

\def \distil  {\vspace{-2pt}\alambic}
\def \oursdis {DeiT\alambic}

\def \oursfix {FixDeiT-B\xspace}

\def \pzo {\phantom{0}} %
\def \alambic {\includegraphics[width=0.02\linewidth]{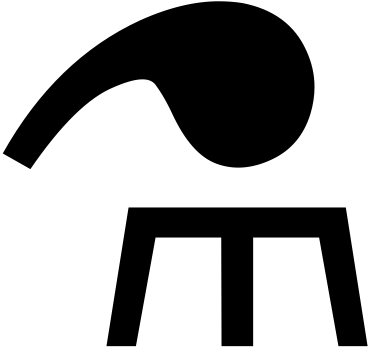}\xspace}

\title{Training data-efficient image transformers \\ \& distillation through attention}
\author{
\begin{minipage}{\linewidth}
\begin{center}
\large Hugo Touvron$^{\star,\dagger}$ \hspace{0.23cm} Matthieu Cord$^{\dagger}$ \hspace{0.23cm} Matthijs Douze$^{\star}$ \\[0.2cm] 
Francisco Massa$^{\star}$ \hspace{0.23cm}
Alexandre Sablayrolles$^{\star}$ \hspace{0.23cm}  Herv\'e J\'egou$^{\star}$ \\[0.5cm]
\scalebox{1.}{$^\star$Facebook AI\hspace{0.6cm} $^\dagger$Sorbonne University}\\[1cm]
\end{center}
\end{minipage}
}

\date{~}

\begin{document}

\maketitle

\begin{abstract}
Recently, neural networks purely based on attention were shown to address image understanding tasks such as image classification. These high-performing vision transformers are pre-trained with hundreds of millions of images using a large infrastructure, thereby limiting their adoption. 

In this work, we produce competitive convolution-free transformers by training on Imagenet only. We train them on a single computer in less than 3 days.
Our reference vision transformer (86M parameters) achieves top-1 accuracy of 83.1\% (single-crop) on ImageNet with no external data.  

More importantly, we introduce a teacher-student strategy specific to  transformers. It relies on a distillation token ensuring that the student learns from the teacher 
through attention. %
We show the interest of this token-based distillation, especially when using a convnet as a teacher. This leads us to report results competitive with convnets for both Imagenet (where we obtain up to 85.2\% accuracy) and when transferring to other tasks. We share our code and models. %
\end{abstract}

\section{Introduction}
\label{sec:introduction}

Convolutional neural networks have been the main design paradigm for image understanding tasks, as initially demonstrated on image classification tasks. One of the ingredient to their success was the availability of a large training set, namely Imagenet~\cite{deng2009imagenet, Russakovsky2015ImageNet12}. 
Motivated by the success of attention-based models in Natural Language Processing~\cite{devlin2018bert,Vaswani2017AttentionIA}, there has been increasing interest in architectures leveraging attention mechanisms within convnets~\cite{Hu2017SENet,Li2019SelectiveKN,zhang2020resnest}. 
More recently several researchers have proposed hybrid architecture transplanting transformer ingredients to convnets to solve vision tasks~\cite{carion2020end,shen2020global}. 

The vision transformer (ViT) introduced by Dosovitskiy et al.~\cite{dosovitskiy2020image} is an architecture directly inherited from Natural Language Processing~\cite{Vaswani2017AttentionIA}, but applied to image classification with raw image patches as input. Their paper presented excellent  results with transformers trained with a large private labelled image dataset (JFT-300M~\cite{sun2017revisiting}, 300 millions images). 
The paper concluded that transformers \emph{``do not generalize well when trained on insufficient amounts of data''}, and the training of these models involved extensive computing resources. 

\begin{figure}[t]
    \centering
    \includegraphics[width=0.72\linewidth]{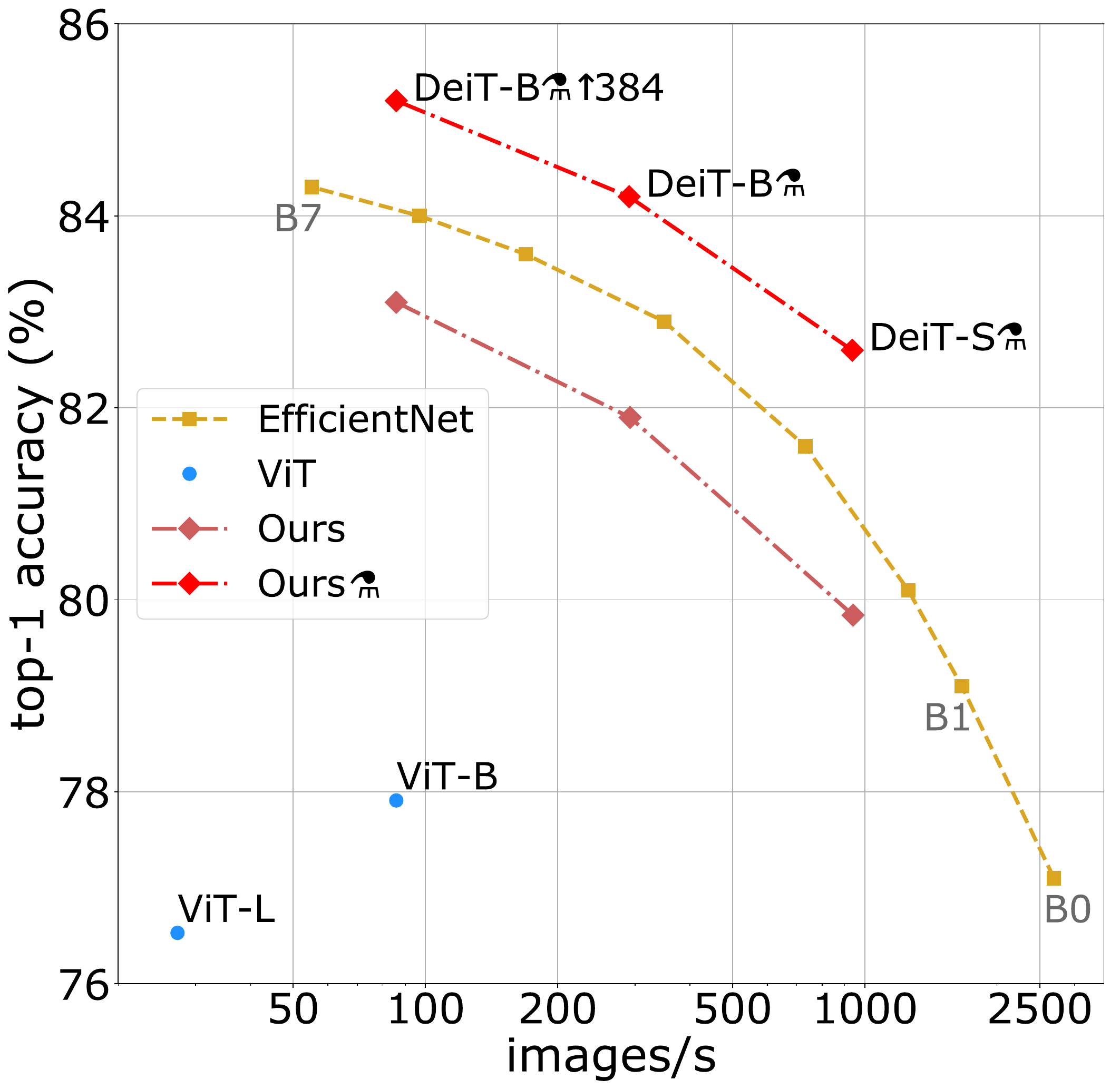}
    \caption{Throughput and accuracy on Imagenet of our methods compared to EfficientNets, trained on Imagenet1k only. The throughput is measured as the number of images processed per second on a V100 GPU. \oursbase is identical to VIT-B, but the training is more adapted to a data-starving regime. It is learned in a few days on one machine. The symbol $\alambic$ refers to models trained with our transformer-specific distillation.
    See Table~\ref{tab:throughput} for details and more models. 
    \label{fig:efficiency}}
\end{figure}

In this paper, we train a vision transformer on a single 8-GPU node in two to three days (53 hours of pre-training, and optionally 20 hours of fine-tuning) that is competitive with convnets having a similar number of parameters and efficiency. It uses Imagenet as the sole training set. 
We build upon the visual transformer architecture from Dosovitskiy et al.~\cite{dosovitskiy2020image} and improvements included in the timm library~\cite{pytorchmodels}. 
With our Data-efficient image Transformers (DeiT), we report large improvements over previous results, see Figure~\ref{fig:efficiency}. 
Our ablation study details the hyper-parameters and key ingredients for a successful training, such as repeated augmentation.  %

We address another question: how to distill these models? 
We introduce a token-based strategy, specific to transformers and denoted by \oursdis, and show that it advantageously replaces the usual distillation. 

In summary, our work makes the following contributions: 
\begin{itemize}
    \item We show that our neural networks that contains no convolutional layer can achieve competitive results against the state of the art on ImageNet with no external data. 
    They are learned on a single node with 4 GPUs in three days\footnote{We can accelerate the learning of the larger model \oursbase by training it on 8 GPUs in two days.}. %
    Our two new models \ourssmall and \ourstiny have fewer parameters and can be seen as the counterpart of ResNet-50 and ResNet-18. 
    \item We introduce a new distillation procedure based on a distillation token, which plays the same role as the class token, except that it aims at reproducing the label estimated by the teacher. Both tokens interact in the transformer through attention. %
    This transformer-specific strategy outperforms  vanilla distillation by a significant margin. 
    \item 
    Interestingly, with our distillation, image  transformers learn more from a convnet than from another transformer with comparable performance. 
    \item Our models pre-learned on Imagenet are competitive when transferred to different downstream tasks such as fine-grained classification, on several popular public benchmarks: CIFAR-10, CIFAR-100, Oxford-102 flowers, Stanford Cars and iNaturalist-18/19. 
\end{itemize}

This paper is organized as follows: we  review related works in Section~\ref{sec:related}, and focus on transformers for image classification in Section~\ref{sec:vit}. %
We introduce our distillation strategy for transformers in Section~\ref{sec:distillation}. 
The experimental section~\ref{sec:experiments} provides analysis and  comparisons against both convnets and recent transformers, as well as a comparative evaluation of our transformer-specific distillation. 
Section~\ref{sec:training} details our training scheme.  
It includes an extensive ablation of our data-efficient training choices, which gives some insight on the key ingredients involved in DeiT. 
We conclude in Section~\ref{sec:conclusion}. 
\section{Related work}
\label{sec:related}

\paragraph{Image Classification} is so core to computer vision that it is often used as a benchmark to measure progress in image understanding. 
Any progress usually translates to improvement in other related tasks such as detection or segmentation. 
Since 2012's AlexNet~\cite{Krizhevsky2012AlexNet}, convnets have dominated this benchmark and have become the de facto standard.
The evolution of the state of the art on the ImageNet dataset~\cite{Russakovsky2015ImageNet12} reflects the progress with convolutional neural network architectures and learning~\cite{Krizhevsky2012AlexNet,Simonyan2015VGG,tan2019efficientnet,Touvron2019FixRes,Touvron2020FixingTT,Xie2019SelftrainingWN}.

Despite several attempts to use transformers for image classification~\cite{chen2020generative}, until now their performance has been inferior to that of convnets.
Nevertheless hybrid architectures that combine convnets and transformers, including the self-attention mechanism, have recently exhibited competitive results in image classification~\cite{wu2020visual}, detection~\cite{carion2020end,Hu2018RelationNF}, video processing~\cite{Sun2019VideoBERTAJ,Wang2018NonlocalNN}, unsupervised object discovery~\cite{Locatello2020ObjectCentricLW}, and unified text-vision tasks~\cite{Chen2020UNITERUI,li2019visualbert,Lu2019ViLBERTPT}.

Recently Vision transformers (ViT)~\cite{dosovitskiy2020image} closed the gap with the state of the art on ImageNet, without using any convolution.
This performance is remarkable since convnet methods for image classification have benefited from years of tuning and optimization~\cite{he2019bag,pytorchmodels}.
Nevertheless, according to this study~\cite{dosovitskiy2020image}, a pre-training phase on a large volume of curated data is required for the learned transformer to be effective. 
In our paper we achieve a strong performance without requiring a large training dataset, i.e., with Imagenet1k only. 
\paragraph{The Transformer architecture,}
introduced by Vaswani et al.~\cite{Vaswani2017AttentionIA} for machine translation are currently the reference model for all natural language processing (NLP) tasks.
Many improvements of convnets for image classification are inspired by transformers.
For example, Squeeze and Excitation~\cite{Hu2017SENet}, Selective Kernel~\cite{Li2019SelectiveKN} and Split-Attention Networks~\cite{zhang2020resnest} exploit mechanism akin to transformers self-attention (SA) mechanism.

\paragraph{Knowledge Distillation} (KD), introduced by Hinton et al.~\cite{Hinton2015DistillingTK}, refers to the training paradigm in which a \emph{student} model leverages ``soft'' labels coming from a strong \emph{teacher} network. 
This is the output vector of the teacher's softmax function rather than just the maximum of scores, wich gives a ``hard'' label.
Such a training improves the performance of the student model (alternatively, it can be regarded as a form of compression of the teacher model into a smaller one -- the student).
On the one hand the teacher's soft labels will have a similar effect to labels smoothing~\cite{Yuan2019RevisitKD}. %
On the other hand as shown by Wei et al.~\cite{Wei2020CircumventingOO} the teacher's supervision takes into account the effects of the data augmentation, which sometimes causes a misalignment between the real label and the image.
For example, let us consider image with a ``cat'' label that represents a large landscape and a small cat in a corner. 
If the cat is no longer on the crop of the data augmentation it implicitly changes the label of the image. 
KD can transfer inductive biases~\cite{abnar2020transferring}  in a soft way in a student model using a teacher model where they would be incorporated in a hard way. 
For example, it may be useful to induce biases due to convolutions in a transformer model by using a convolutional model as teacher. 
In our paper we study the distillation of a transformer student by either a convnet or a transformer teacher. We introduce a new distillation procedure specific to transformers and show its superiority. 

\section{Vision transformer: overview} 
\label{sec:vit} 

In this section, we briefly recall preliminaries associated with the vision transformer~\cite{dosovitskiy2020image,Vaswani2017AttentionIA}, and further discuss positional encoding and resolution. 


\paragraph{Multi-head Self Attention layers (MSA).}
The attention mechanism is based on a trainable associative memory with (key, value) vector pairs. 
A \emph{query} vector $q\in\mathbb{R}^d$ is matched against a set of $k$ \emph{key} vectors (packed together into a matrix $K\in\mathbb{R}^{k\times d}$) using inner products. 
These inner products are then scaled and normalized with a softmax function to obtain $k$ weights. 
The output of the attention is the weighted sum of a set of $k$ \emph{value} vectors (packed into $V\in\mathbb{R}^{k\times d}$). 
For a sequence of $N$ query vectors (packed into $Q\in\mathbb{R}^{N\times d}$), it produces an output matrix (of size $N\times d$): 
\begin{equation}
    \mathrm{Attention}(Q, K, V) = \mathrm{Softmax}(Q K^\top/\sqrt{d}) V,
\end{equation}
where the $\mathrm{Softmax}$ function is applied over each row of the input matrix and the $\sqrt{d}$ term provides appropriate normalization.\\
In~\cite{Vaswani2017AttentionIA}, a Self-attention layer is proposed.  Query, key and values matrices are themselves computed from a sequence of $N$ input vectors (packed into $X\in \mathbb{R}^{N\times D}$): 
$Q=XW_\mathrm{Q}$, $K=XW_\mathrm{K}$, $V=XW_\mathrm{V}$, using linear transformations $W_\mathrm{Q},W_\mathrm{K},W_\mathrm{V}$ with the constraint $k=N$, meaning that the attention is in between all the input vectors. \\
Finally, {Multi-head} self-attention layer (MSA) is defined by considering $h$ attention ``heads'', \textit{ie} $h$ self-attention functions  applied to the input. 
Each head provides a sequence of size $N\times d$. 
These $h$ sequences are rearranged into a $N \times dh$ sequence that is reprojected by a linear layer into $N \times D$. 

\paragraph{Transformer block for images.}
To get a full transformer block as in~\cite{Vaswani2017AttentionIA},  we add a Feed-Forward Network (FFN) on top of the MSA layer. This FFN is composed of two linear layers separated by a GeLu activation~\cite{Hendrycks2016GaussianEL}.
The first linear layer expands the dimension from $D$ to $4D$, and the second layer reduces the dimension from $4D$ back to $D$.
Both MSA and FFN are operating as residual operators thank to skip-connections, and with a layer normalization~\cite{ba2016layer}.

In order to get a transformer to process images, our work builds upon the ViT model~\cite{dosovitskiy2020image}.  It is a simple and elegant architecture that processes input images as if they were a sequence of input tokens. The fixed-size input RGB image is decomposed into a batch of $N$ patches of a fixed size of $16\times 16$ pixels ($N=14\times 14$). Each patch is projected with a linear layer that conserves its overall dimension $3 \times 16 \times 16=768$.\\
The transformer block described above is invariant to the order of the patch embeddings, and thus does not consider their relative position.
The positional information is incorporated as fixed~\cite{Vaswani2017AttentionIA} or trainable~\cite{ghering2017convseq2seq} positional embeddings. 
They are added before the first transformer block to the patch tokens, which are then fed to the stack of transformer blocks.

\paragraph{The class token} is a trainable vector, appended to the patch tokens before the first layer, that goes through the transformer layers, and is then projected with a linear layer to predict the class.
This class token is inherited from NLP~\cite{devlin2018bert}, and departs from the typical pooling layers used in computer vision to predict the class. 
The transformer thus process batches of $(N + 1)$ tokens of dimension $D$, of which only the class vector is used to predict the output.
This architecture forces the self-attention to spread information between the patch tokens and the class token: at training time the supervision signal comes only from the class embedding, while the patch tokens are the model's only variable input.

\paragraph{Fixing the positional encoding across resolutions. }
\label{sec:position_encoding}

Touvron et al.~\cite{Touvron2019FixRes} show that it is  desirable to use a lower training resolution and fine-tune the network at the larger resolution. %
This speeds up the full training and improves the accuracy under prevailing data augmentation schemes. 
When increasing the resolution of an input image, we keep the patch size the same, 
therefore the number $N$ of input patches does change.
Due to the architecture of transformer blocks and the class token, the model and classifier do not need to be modified to process more tokens.
In contrast, one needs to adapt the positional embeddings, because there are $N$ of them, one for each patch. 
Dosovitskiy et al.~\cite{dosovitskiy2020image}  interpolate the positional encoding when changing the resolution and demonstrate that this method works with the subsequent fine-tuning stage. 

\section{Distillation through attention}
\label{sec:distillation} 

In this section we assume we have access to a strong image classifier as a teacher model.
It could be a convnet, or a mixture of classifiers.
We address the question of how to learn a transformer by exploiting this teacher.
As we will see in Section~\ref{sec:experiments} by comparing the trade-off between accuracy and image throughput, 
it can be beneficial to replace a convolutional neural network by a transformer. 
This section covers two axes of distillation: hard distillation versus soft distillation, and classical distillation versus the distillation token.

\begin{figure}[t]
\centering \includegraphics[width=0.5\linewidth]{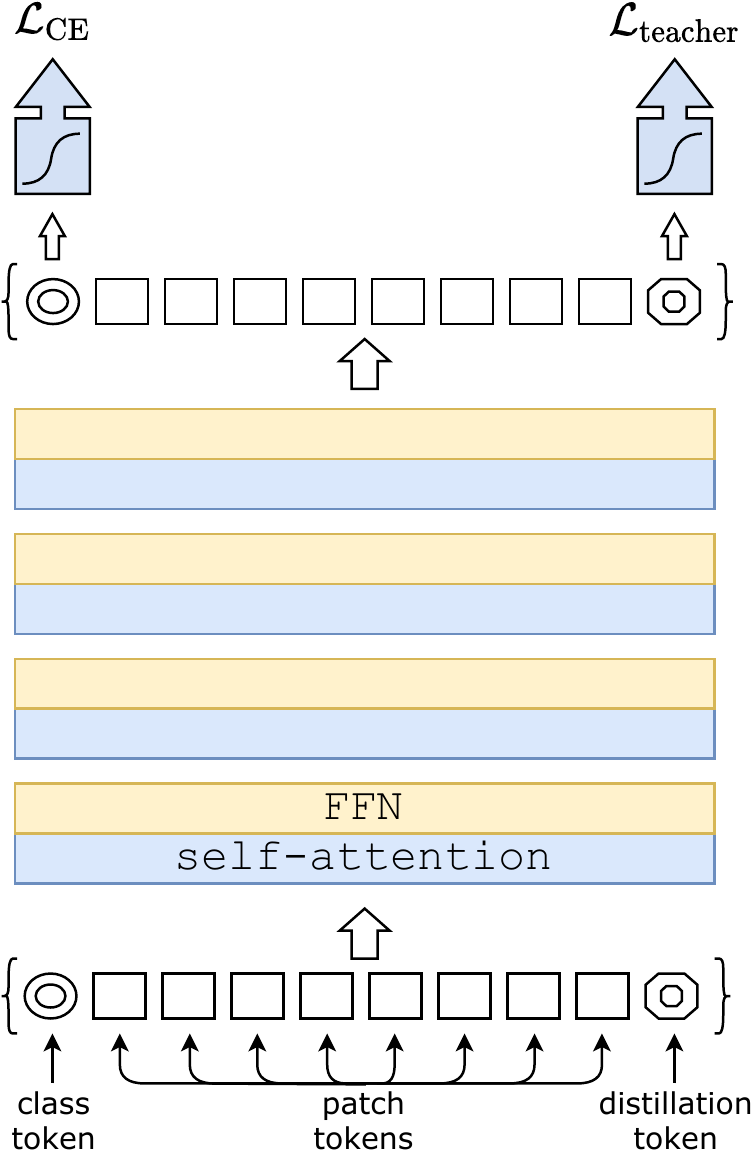}
\caption{Our distillation procedure: 
we simply include a new \emph{distillation token}. It interacts with the class and patch tokens through the self-attention layers. This distillation token is employed in a similar fashion as the class token, except that on output of the network its objective is to reproduce the (hard) label predicted by the teacher, instead of true label. Both the class and distillation tokens input to the transformers are learned by back-propagation. 
\label{fig:distillation}}
\end{figure}

\paragraph{Soft distillation}  
\cite{Hinton2015DistillingTK,Wei2020CircumventingOO} minimizes the Kullback-Leibler divergence between the softmax of the teacher and the softmax of the student model. 

Let $Z_\mathrm{t}$ be the logits of the teacher model, $Z_\mathrm{s}$ the logits of the student model.
We denote by $\tau$ the temperature for the distillation, $\lambda$ the coefficient balancing the Kullback–Leibler divergence loss ($\mathrm{KL}$) and the cross-entropy ($\mathcal{L}_\mathrm{CE}$) on ground truth labels $y$, and $\psi$ the softmax function.
The distillation objective is

\begin{equation}
    \mathcal{L}_\mathrm{global} = (1-\lambda) \mathcal{L}_\mathrm{CE}(\psi(Z_\mathrm{s}),y) + \lambda  \tau^2 \mathrm{KL}(\psi(Z_\mathrm{s}/\tau),\psi(Z_\mathrm{t}/\tau)).
\end{equation}

\paragraph{Hard-label distillation.} %
We introduce a variant of distillation where we take the hard decision of the teacher as a true label.
Let $y_\mathrm{t}=\mathrm{argmax}_c Z_\mathrm{t}(c)$ be the hard decision of the teacher, the objective associated with 
this hard-label distillation is: %
\begin{equation}
    \mathcal{L}_\mathrm{global}^\mathrm{hard Distill} =  \frac{1}{2}\mathcal{L}_\mathrm{CE}(\psi(Z_s),y) +  \frac{1}{2}\mathcal{L}_\mathrm{CE}(\psi(Z_s),y_\mathrm{t}).
\end{equation}

For a given image, the hard label associated with the teacher may change depending on the specific data augmentation. 
We will see that this choice is better than the traditional one, while being parameter-free and conceptually simpler: The teacher prediction $y_\mathrm{t}$ plays the same role as the true label $y$.

Note also that the hard labels can also be converted into soft labels with label smoothing~\cite{Szegedy2016RethinkingTI}, where the true label is considered to have a probability of $1-\varepsilon$, and the remaining $\varepsilon$ is shared across the remaining classes. We fix this parameter to $\varepsilon=0.1$ in our all experiments that use true labels.  

\paragraph{Distillation token.} 
We now focus on our proposal, which is illustrated in Figure~\ref{fig:distillation}. 
We add a new token, the distillation token, to the initial embeddings (patches and class token).
Our distillation token is used similarly as the class token: it interacts with other embeddings through self-attention, and is output by the network after the last layer.
Its target objective is given by the distillation component of the loss. 
The distillation embedding allows our model to learn from the output of the teacher, as in a regular distillation, while remaining complementary to the class embedding.

Interestingly, we observe that the learned class and distillation tokens converge towards different vectors: the average cosine similarity between these tokens equal to 0.06. As the class and distillation embeddings are computed at each layer, they gradually become more similar through the network, all the way through the last layer at which their similarity is high (cos=0.93), but still lower than 1. This is expected since as they aim at producing targets that are similar but not identical. 

We verified that our distillation token adds something to the model, compared to simply adding an additional class token associated with the same target label:  instead of a teacher pseudo-label, we experimented with a transformer with two class tokens. Even if we initialize them randomly and independently, during training they converge towards the same vector (cos=0.999), and the output embedding are also quasi-identical.  This additional class token does not bring anything to the classification performance. 
In contrast, our distillation strategy provides a significant improvement over a vanilla distillation baseline, as validated by our experiments in Section~\ref{sec:distillation_results}.  

\paragraph{Fine-tuning with distillation.} 
We use both the true label and teacher prediction during the fine-tuning stage at higher resolution. We use a teacher with the same target resolution, typically obtained from the lower-resolution teacher by the method of Touvron et al~\cite{Touvron2019FixRes}. We have also tested with true labels only but this reduces the benefit of the teacher and leads to a lower performance.

\paragraph{Classification with our approach: joint classifiers.} 
At test time,  both the class or the distillation embeddings produced by the transformer are associated with linear classifiers and able to infer the image label. Yet our referent method is the late fusion of these two separate heads, for which we add the softmax output by the two classifiers to make the prediction. We evaluate these three options in Section~\ref{sec:experiments}. 

\section{Experiments}
\label{sec:experiments}

This section presents a few analytical experiments and results. We first discuss our distillation strategy. Then we comparatively analyze the efficiency and accuracy of convnets and vision transformers. 

\subsection{Transformer models}
As mentioned earlier, our architecture design is identical to the one proposed by Dosovitskiy et al.~\cite{dosovitskiy2020image} with no convolutions. 
Our only differences are the training strategies, and the distillation token. Also we do not use a MLP head for the pre-training but only a linear classifier. 
To avoid any confusion, we refer to the results obtained in the prior work by ViT, and prefix ours by \ours. If not specified, \ours refers to our referent model \oursbase, which has the same architecture as ViT-B. 
When we fine-tune \ours at a larger resolution, we append the resulting operating resolution at the end, e.g, \oursbaseup384.  
Last, when using our distillation procedure, we identify it with an alembic sign as \oursdis. 

The parameters of ViT-B (and therefore of \oursbase) are fixed as  $D=768$, $h=12$ and $d=D/h=64$.
We introduce two smaller models, namely \ourssmall and \ourstiny, for which we change the number of heads, keeping $d$ fixed. Table~\ref{tab:models} summarizes the models that we consider in our paper.

\begin{table}[t]
\caption{Variants of our DeiT architecture.
The larger model, DeiT-B, has the same architecture as the ViT-B~\cite{dosovitskiy2020image}.
The only parameters that vary across models are the embedding dimension and the number of heads, and we keep the dimension per head constant (equal to 64). 
Smaller models have a lower parameter count, and a faster throughput. The throughput is measured for images at resolution 224$\times$224. 
\label{tab:models}}
\smallskip
\centering
\scalebox{0.8}
{
\begin{tabular}{l|cccccccc}
\toprule
Model       & ViT model         & embedding  & \#heads & \#layers & \#params & training & throughput  \\
            &                   & dimension  &         &           &          & resolution & (im/sec)\\
\midrule
\ourstiny   &  N/A  &  \pzo192  & \pzo3 & 12 & \pzo\pzo5M & 224 & 2536 \\
\ourssmall  &  N/A  &  \pzo384 & \pzo6 & 12 & \pzo22M& 224& \pzo940 \\
\oursbase   &  ViT-B & \pzo768   & 12 & 12 & \pzo86M & 224 & \pzo292\\
\bottomrule
\end{tabular}}
\end{table}

\subsection{Distillation}
\label{sec:distillation_results}

Our distillation method produces a vision transformer that becomes on par with the best convnets in terms of the trade-off between accuracy and throughput, see Table~\ref{tab:throughput}. Interestingly, the distilled model outperforms its teacher in terms of the trade-off between accuracy and throughput. Our best model on ImageNet-1k is 85.2\% top-1 accuracy outperforms the best Vit-B model pre-trained on JFT-300M at resolution 384 (84.15\%). For reference, the current state of the art of 88.55\% achieved with extra training data was obtained by the ViT-H model (600M parameters) trained on JFT-300M at resolution 512. 
Hereafter we provide several analysis and observations. 

\paragraph{Convnets teachers.}

We have observed that using a convnet teacher gives better performance than using a transformer.
Table~\ref{tab:teacher} compares distillation results with different teacher architectures.
The fact that the convnet is a better teacher is probably due to the inductive bias inherited by  the transformers through distillation, as explained in Abnar et al.~\cite{abnar2020transferring}.
In all of our subsequent distillation experiments the default teacher is a RegNetY-16GF~\cite{Radosavovic2020RegNet} (84M parameters) that we trained with the same data and same data-augmentation as \ours.
This teacher reaches $82.9\%$ top-1 accuracy on ImageNet.

\begin{table}[t]
\caption{We compare on ImageNet~\cite{Russakovsky2015ImageNet12} the performance (top-1 acc., \%) of the student as a function of the teacher model used for distillation. %
\label{tab:teacher}
\smallskip
}
\centering
\scalebox{0.9}
{
\begin{tabular}{lc|cc}
\toprule
\multicolumn{2}{c}{Teacher}   & \multicolumn{2}{c}{Student: \oursbase\distil} \\
 Models & acc. & pretrain & $\uparrow$384\\
\midrule
\oursbase  & 81.8 & 81.9 & 83.1\\
\midrule
RegNetY-4GF & 80.0 & 82.7 & 83.6\\
RegNetY-8GF & 81.7 & 82.7 &  83.8\\
RegNetY-12GF & 82.4 & 83.1 & 84.1\\
RegNetY-16GF & 82.9 &  83.1 & 84.2\\
\bottomrule
\end{tabular}}
\end{table}

\paragraph{Comparison of distillation methods.} 

We compare the performance of different distillation strategies in Table~\ref{tab:distillation}. 
Hard distillation significantly outperforms soft distillation for transformers, even when using only a class token: hard distillation reaches 83.0\% at resolution 224$\times$224, compared to the soft distillation accuracy of 81.8\%.
Our distillation strategy from Section~\ref{sec:distillation} further improves the performance, showing that the two tokens provide complementary information useful for classification: the classifier %
on the two tokens is significantly better than the independent class and distillation classifiers, which by themselves already outperform the distillation baseline. 

The distillation token gives slightly better results than the class token. 
It is also more correlated to the convnets prediction. This difference in performance is probably due to the fact that it benefits more from the inductive bias of convnets.
We give more details and an analysis in the next paragraph.
The distillation token has an undeniable advantage for the  initial training. 

\paragraph{Agreement with the teacher \& inductive bias?}
As discussed above, the architecture of the teacher has an important impact.  Does it inherit existing inductive bias that would facilitate the training?  %
While we believe it difficult to formally answer this question, we analyze in  
Table~\ref{tab:agreement} the decision agreement between the convnet teacher, our image transformer \ours learned from labels only, and our transformer \oursdis. 

Our distilled model is more correlated to the convnet than with a transformer learned from scratch. As to be expected, the classifier associated with the distillation embedding is closer to the convnet that the one associated with the class embedding, and conversely the one associated with the class embedding is more similar to \ours learned without distillation. Unsurprisingly, the joint class+distil classifier offers a middle ground.  

\begin{table}[t]
\caption{Distillation experiments on Imagenet with \ours, 300 epochs of pre-training. 
We report the results for our new distillation method in the last three rows. 
We separately report the performance when classifying with only one of the class or distillation embeddings, and then with a classifier taking both of them as input.  \label{tab:distillation}
In the last row (class+distillation), the result %
correspond to the late fusion of the class and distillation classifiers. %
}
\smallskip
\centering
\scalebox{0.88}
{
\begin{tabular}{l|cc|cccc}
\toprule
                              &  \multicolumn{2}{c}{Supervision}  &  \multicolumn{4}{c}{ImageNet top-1 (\%)} \\
 method $\downarrow$          & label & teacher        & Ti 224  & S 224 & B 224 & B$\uparrow$384 \\
\midrule
 \ours -- no distillation    & \cmark &  \xmark  & 72.2 & 79.8    & 81.8  & 83.1 \\
 \ours -- usual distillation & \xmark & soft & 72.2 & 79.8    & 81.8  &  83.2    \\ 
 \ours -- hard  distillation &  \xmark& hard & 74.3 & 80.9    & 83.0  & 84.0 \\ 
\midrule 
\oursdis: class embedding   & \cmark & hard & 73.9    & 80.9  & 83.0  &  84.2\\
\oursdis: distil. embedding & \cmark & hard & 74.6      &  81.1  & 83.1 & 84.4 \\ 
\oursdis: class+distillation& \cmark & hard & 74.5      & 81.2 & 83.4  &  84.5 \\
\bottomrule
\end{tabular}}
\end{table}

\begin{figure}[t]
    \centering
    \includegraphics[width=0.75\linewidth]{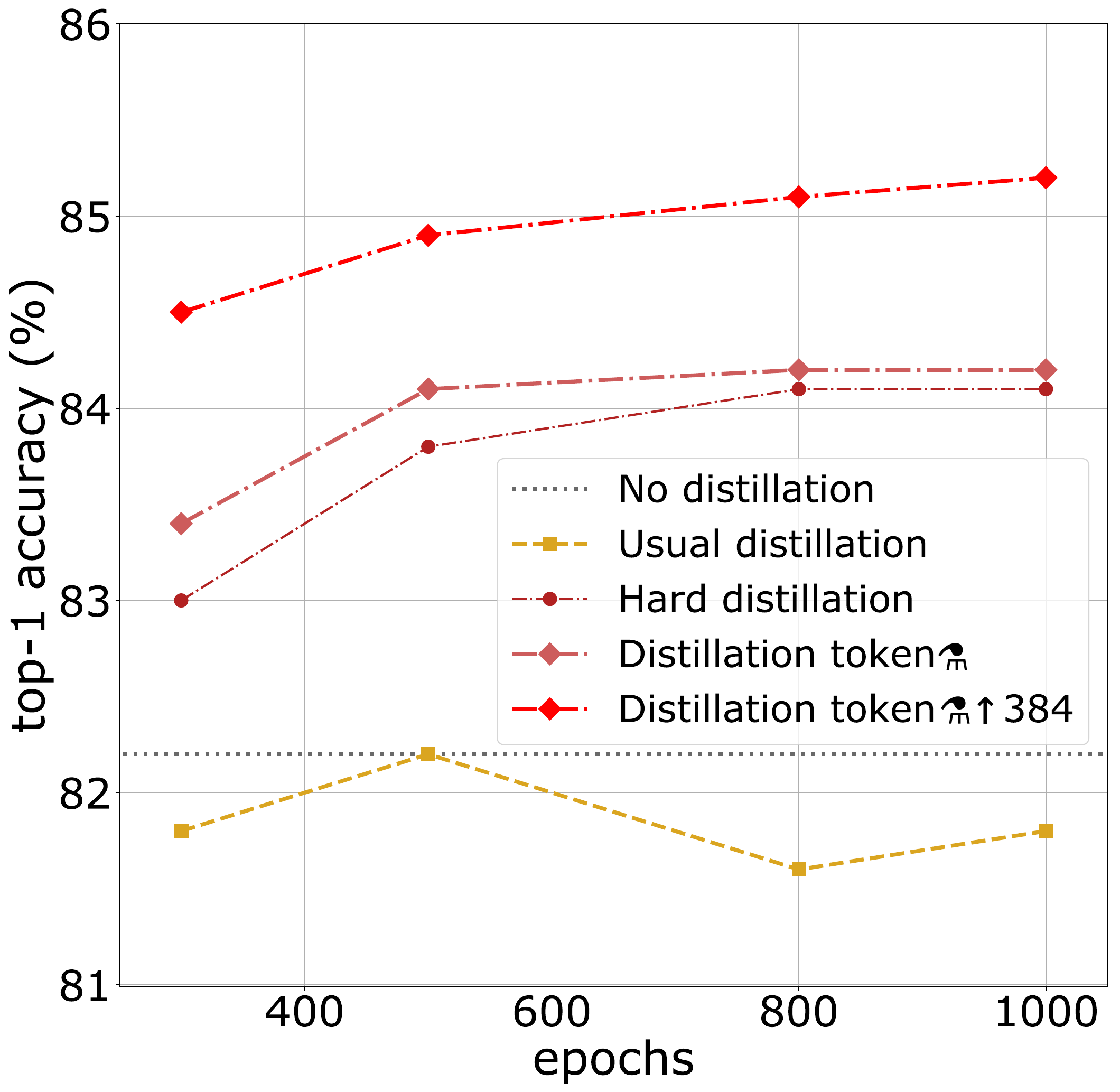}
    \caption{Distillation on ImageNet~\cite{Russakovsky2015ImageNet12} with \ours-B: performance as a function of the number of training epochs. We provide the performance without distillation (horizontal dotted line) as it saturates after 400 epochs. 
    \label{fig:distillation_epochs}}
\end{figure}

\paragraph{Number of epochs.} %
Increasing the number of epochs significantly improves the performance of training with distillation, see Figure~\ref{fig:distillation_epochs}. With 300 epochs, our distilled network \oursbasedis is already better than \oursbase. But while for the latter the performance saturates with longer schedules, our distilled network clearly benefits from a longer training time. 

\begin{table}[t]
\caption{Disagreement analysis between convnet, image transformers and distillated transformers: We report the fraction of sample classified differently for all classifier pairs, i.e., the rate of different decisions. We include two models without distillation (a RegNetY and DeiT-B), so that we can compare how our distilled models and classification heads are correlated to these teachers. \smallskip  
\label{tab:agreement}}
\centering
\scalebox{0.86}
{
\begin{tabular}{l|c|cc|ccc}
\toprule
             & \multirow{2}{*}{groundtruth}  &  \multicolumn{2}{c}{no distillation}  & \multicolumn{3}{c}{\oursdis student (of the convnet)} \\ 
             &              & convnet & \ours &  class  &  distillation &  \oursdis \\
\midrule 
groundtruth                 & 0.000        & 0.171   & 0.182    & 0.170  &  0.169 &  0.166      \\
convnet (RegNetY)           & 0.171        & 0.000   & 0.133    & 0.112  & 0.100 &   0.102     \\
\ours                       & 0.182        & 0.133   &  0.000   &   0.109            & 0.110            &  0.107      \\
\midrule
\oursdis -- class only   &  0.170 &  0.112 &  0.109       & 0.000         & 0.050 &   0.033     \\
\oursdis -- distil. only & 0.169 &  0.100 &   0.110      & 0.050  &  0.000     &  0.019      \\
\oursdis -- class+distil.&  0.166 &  0.102 &  0.107       &      0.033         &   0.019        &  0.000  \\
\bottomrule
\end{tabular}}
\end{table}

\subsection{Efficiency vs accuracy: a comparative study with convnets}

In the literature, the image classificaton methods are often compared as a compromise between accuracy and another criterion, such as FLOPs, number of parameters, size of the network, etc. 

We focus  in Figure~\ref{fig:efficiency} on the tradeoff between the throughput (images processed per second) and the top-1 classification accuracy on ImageNet. 
We focus on the popular state-of-the-art EfficientNet convnet, which has benefited from years of research on convnets and was optimized by architecture search on the ImageNet validation set. %

Our method \ours is slightly below EfficientNet, which shows that we have almost closed the gap between vision transformers and convnets when training with Imagenet only. These results are a major improvement (\textbf{+6.3\%} top-1 in a comparable setting) over previous ViT models trained on Imagenet1k only~\cite{dosovitskiy2020image}. 
Furthermore, when \ours benefits from the distillation from a relatively weaker RegNetY to produce \oursdis, it outperforms EfficientNet. It also outperforms by 1\% (top-1 acc.) the Vit-B model pre-trained on JFT300M at resolution 384 (85.2\% vs  84.15\%), while being significantly faster to train. 

Table~\ref{tab:throughput} reports the numerical results in more details and additional evaluations on ImageNet V2 and ImageNet Real, that have a test set distinct from the ImageNet validation, which reduces overfitting on the validation set. Our results show that \oursbasedis and \oursbasedisup384  outperform, by some margin, the state of the art on the trade-off between accuracy and inference time on GPU. 

\begin{table}[p]
    \centering
    \scalebox{0.85}
    {\small 
    \begin{tabular}{@{\ }l|r|c@{\ }c|c@{\ }|c@{\ }|c@{\ }}
    \toprule
            &              &     image       & throughput & \multicolumn{1}{c}{ImNet} & \multicolumn{1}{|c|}{Real}& \multicolumn{1}{c}{V2}\\ 
    Network & \#param. & size & (image/s) &  top-1 &  top-1 &  top-1\\
    \toprule
    \multicolumn{7}{c}{Convnets}\\
    \midrule
    ResNet-18~\cite{He2016ResNet} & 12M & $224^{2}$ & 4458.4 &  69.8  & 77.3 & 57.1\\
    ResNet-50~\cite{He2016ResNet} & 25M & $224^{2}$ & 1226.1 & 76.2 & 82.5 & 63.3\\
    ResNet-101~\cite{He2016ResNet}& 45M & $224^{2}$ & \pzo753.6 & 77.4 & 83.7 & 65.7\\
    ResNet-152~\cite{He2016ResNet} & 60M & $224^{2}$ & \pzo526.4 & 78.3& 84.1 & 67.0 \\
    \midrule
    RegNetY-4GF~\cite{Radosavovic2020RegNet}$\star$  &  21M & $224^{2}$ & 1156.7 & 80.0 & 86.4 & 69.4 \\
    RegNetY-8GF~\cite{Radosavovic2020RegNet}$\star$  &  39M & $224^{2}$ & \pzo591.6 & 81.7 & 87.4  & 70.8 \\ 
    RegNetY-16GF~\cite{Radosavovic2020RegNet}$\star$ &  84M & $224^{2}$ & \pzo334.7 & 82.9 & 88.1 & 72.4 \\
    \midrule
    EfficientNet-B0~\cite{tan2019efficientnet} & 5M & $224^{2}$ & 2694.3 & 77.1 & 83.5 & 64.3\\
    EfficientNet-B1~\cite{tan2019efficientnet} & 8M & $240^{2}$ & 1662.5 & 79.1  & 84.9 & 66.9 \\
    EfficientNet-B2~\cite{tan2019efficientnet} & 9M & $260^{2}$ & 1255.7  & 80.1 & 85.9 & 68.8 \\
    EfficientNet-B3~\cite{tan2019efficientnet} &12M &$300^{2}$  & \pzo732.1 & 81.6  & 86.8 & 70.6 \\
    EfficientNet-B4~\cite{tan2019efficientnet} & 19M & $380^{2}$ & \pzo349.4 & 82.9  & 88.0& 72.3 \\
    EfficientNet-B5~\cite{tan2019efficientnet} & 30M & $456^{2}$ & \pzo169.1 & 83.6  & 88.3 & 73.6 \\
    EfficientNet-B6~\cite{tan2019efficientnet} & 43M & $528^{2}$ & \pzo\pzo96.9 & 84.0 & 88.8 & 73.9 \\
    EfficientNet-B7~\cite{tan2019efficientnet} & 66M & $600^{2}$ & \pzo\pzo55.1 & 84.3 & \_& \_ \\
    \midrule
    EfficientNet-B5 RA~\cite{Cubuk2019RandAugmentPA} & 30M & $456^{2}$ & \pzo\pzo96.9 & 83.7  & \_& \_ \\
    EfficientNet-B7 RA~\cite{Cubuk2019RandAugmentPA} & 66M & $600^{2}$ & \pzo\pzo55.1 & 84.7 & \_& \_ \\
    \midrule
    KDforAA-B8 & 87M & $800^{2}$ & \pzo\pzo25.2 & 85.8 & \_ & \_ \\
    \toprule
    \multicolumn{7}{c}{Transformers}\\
    \midrule
    ViT-B/16~\cite{dosovitskiy2020image} & 86M  & $384^{2}$ & \pzo\pzo85.9 & 77.9 & 83.6 & \_ \\
    ViT-L/16~\cite{dosovitskiy2020image} & 307M & $384^{2}$ & \pzo\pzo27.3 & 76.5  & 82.2 & \_ \\
    \midrule
    \ourstiny & 5M & $224^{2}$ & 2536.5 & 72.2  & 80.1  &  60.4\\
    \ourssmall & 22M & $224^{2}$ & \pzo940.4 & 79.8 & 85.7  & 68.5 \\
    \oursbase & 86M & $224^{2}$ & \pzo292.3 & 81.8  & 86.7 & 71.5\\
    \midrule
    \oursbaseup384 & 86M & $384^{2}$ & \pzo\pzo85.9 & 83.1 & 87.7 & 72.4\\
    
    \midrule
    \ourstinydis & 6M & $224^{2}$ & 2529.5 & 74.5  &  82.1  & 62.9 \\
    \ourssmalldis & 22M & $224^{2}$ & \pzo936.2 & 81.2 & 86.8  & 70.0  \\

    \oursbasedis & 87M &  $224^{2}$ & \pzo290.9 & 83.4 & 88.3 & 73.2 \\
    \midrule
    \ourstinydis / 1000 epochs  & 6M &  $224^{2}$ & 2529.5 & 76.6  & 83.9 & 65.4 \\
    \ourssmalldis / 1000 epochs & 22M &  $224^{2}$ & \pzo936.2 & 82.6 & 87.8 & 71.7\\
    \oursbasedis / 1000 epochs  & 87M &  $224^{2}$ & \pzo290.9 & 84.2 & 88.7 & 73.9 \\

    \midrule
    \oursbasedisup384 & 87M & $384^{2}$ & \pzo\pzo85.8 & 84.5  & 89.0 & 74.8 \\
    \midrule
    \oursbasedisup384 / 1000 epochs & 87M & $384^{2}$ & \pzo\pzo85.8 & 85.2  & 89.3 & 75.2 \\
    \bottomrule
    \end{tabular}}
    \caption{Throughput on and accuracy on Imagenet~\cite{Russakovsky2015ImageNet12}, Imagenet Real~\cite{Beyer2020ImageNetReal} and Imagenet V2 matched frequency~\cite{Recht2019ImageNetv2} of \ours and of several state-of-the-art convnets, for models trained with no external data.  The throughput is measured as the number of images that we can process per second  on one 16GB V100 GPU. For each model we take the largest possible batch size for the usual resolution of the model and calculate the average time over 30 runs to process that batch. With that we calculate the number of images processed per second. Throughput can vary according to the implementation: for a direct comparison and in order to be as fair as possible, we use for each model the definition in the same GitHub~\cite{pytorchmodels} repository. 
    \newline 
    $\star:$ \textit{Regnet optimized with a similar optimization procedure as ours, which boosts the results. These networks serve as teachers when we use our distillation strategy. }
    \label{tab:throughput}}
\end{table}

\subsection{Transfer learning: Performance on downstream tasks}

Although \ours perform very well on ImageNet it is important to evaluate them on other datasets with transfer learning in order to measure the power of generalization of \ours.
We evaluated this on transfer learning  tasks by fine-tuning on the datasets in Table~\ref{tab:dataset}.
Table~\ref{tab:sota} compares \ours transfer learning results  to those of ViT~\cite{dosovitskiy2020image} and state of the art convolutional architectures~\cite{tan2019efficientnet}.
\ours is on par with competitive convnet models, which is in line with our previous conclusion on ImageNet.

\begin{table}[t]
\caption{Datasets used for our different tasks.  \label{tab:dataset}}
\smallskip
\centering
{\small
\begin{tabular}{l|rrr}
\toprule
Dataset & Train size & Test size & \#classes   \\
\midrule
ImageNet \cite{Russakovsky2015ImageNet12}  & 1,281,167 & 50,000 & 1000  \\ 
iNaturalist 2018~\cite{Horn2018INaturalist}& 437,513   & 24,426 & 8,142 \\ 
iNaturalist 2019~\cite{Horn2019INaturalist}& 265,240   & 3,003  & 1,010  \\ 
Flowers-102~\cite{Nilsback08}& 2,040 & 6,149 & 102  \\ 
Stanford Cars~\cite{Cars2013}& 8,144 & 8,041 & 196  \\  
CIFAR-100~\cite{Krizhevsky2009LearningML}  & 50,000    & 10,000 & 100   \\ 
CIFAR-10~\cite{Krizhevsky2009LearningML}  & 50,000    & 10,000 & 10   \\ 
\bottomrule
\end{tabular}}
\vspace{-8pt}
\end{table}

\begin{table}[t]
    \caption{We compare Transformers based models on different transfer learning task with ImageNet pre-training. We also report results with convolutional architectures for reference. 
    \label{tab:sota}}
    \smallskip
    \centering
    \scalebox{0.68}{
    \begin{tabular}{l|c|cccccc|r}
    \toprule
    Model                                      & ImageNet & CIFAR-10 & CIFAR-100  & Flowers & Cars & iNat-18 & iNat-19 & im/sec \\
    \midrule                                                                          
    Grafit ResNet-50~\cite{Touvron2020GrafitLF} & 79.6 & \_   & \_ & 98.2 & 92.5 & 69.8 & 75.9 & 1226.1\\
    Grafit RegNetY-8GF~\cite{Touvron2020GrafitLF} & \_ & \_   & \_ & 99.0 & 94.0 & 76.8 & 80.0 & 591.6\\
    ResNet-152~\cite{Chu2020FeatureSA} & \_ & \_ & \_ & \_ & \_ & 69.1 & \_ & 526.3\\
    EfficientNet-B7~\cite{tan2019efficientnet}  & 84.3 & 98.9 & 91.7  & 98.8 & 94.7 & \_ & \_ & 55.1\\
    \midrule                                                                          
    ViT-B/32~\cite{dosovitskiy2020image}        & 73.4 & 97.8 & 86.3 & 85.4 & \_   & \_ & \_ & 394.5\\
    ViT-B/16~\cite{dosovitskiy2020image}        & 77.9 & 98.1 & 87.1 & 89.5 & \_   & \_ & \_ & 85.9 \\
    ViT-L/32~\cite{dosovitskiy2020image}        & 71.2 & 97.9 & 87.1 & 86.4 & \_   & \_ & \_ & 124.1\\
    ViT-L/16~\cite{dosovitskiy2020image}        & 76.5 & 97.9 & 86.4 & 89.7 & \_   & \_ & \_ & 27.3 \\
\midrule
    \oursbase                                     & 81.8 & 99.1 &  90.8  & 98.4 &  92.1    & 73.2 & 77.7 &  292.3 \\
    \oursbaseup384    & 83.1 & 99.1 &  90.8  & 98.5 &  93.3    &  79.5 &  81.4    &  \pzo85.9 \\
    \oursbasedis   & 83.4 & 99.1  &  91.3  & 98.8 & 92.9 & 73.7  & 78.4 &   290.9\\
    \oursbasedisup384 & 84.4 & 99.2 & 91.4   & 98.9 & 93.9 & 80.1 & 83.0 &  \pzo85.9  \\
    \bottomrule
    \end{tabular}}
\end{table}

\paragraph{Comparison vs training from scratch.}
We investigate the performance when training from scratch on a small dataset, without Imagenet pre-training. We get the following results on the small CIFAR-10, which is small both w.r.t. the number of images and labels:
\begin{center}
    {\small 
    \begin{tabular}{c|c|c|c}
    \toprule
    Method  & RegNetY-16GF & \ours-B & \oursbasedis \\
   Top-1 & 98.0 & 97.5 & 98.5 \\ 
    \bottomrule
    \end{tabular}}
\end{center}

For this experiment, we tried we get as close as possible to the Imagenet pre-training counterpart, meaning that (1) we consider longer training schedules (up to 7200 epochs, which corresponds to 300 Imagenet epochs) so that the network has been fed a comparable number of images in total; (2) we re-scale images to $224\times224$ to ensure that we have the same augmentation. 
The results are not as good as with Imagenet pre-training (98.5\% vs 99.1\%), which is expected since the network has seen a much lower diversity. However they show that it is possible to learn a reasonable transformer on CIFAR-10 only.

\section{Training details \& ablation}
\label{sec:training}

\begin{table*}[t]
    \centering
    \scalebox{0.85}
    {\small
    \begin{tabular}{c|cc|cccc|ccccc|cc}

\multicolumn{12}{c}{~} & \multicolumn{2}{c}{top-1 accuracy} \\  
    \cmidrule(lr){13-14} 
    
    Ablation on $\downarrow$
    & \rotatebox{90}{Pre-training}
    & \rotatebox{90}{Fine-tuning}
    & \rotatebox{90}{Rand-Augment}
    & \rotatebox{90}{AutoAug}
    & \rotatebox{90}{Mixup}
    & \rotatebox{90}{CutMix}
    
    & \rotatebox{90}{Erasing}
    & \rotatebox{90}{Stoch. Depth}
    & \rotatebox{90}{Repeated Aug.} 
    & \rotatebox{90}{Dropout} 
    & \rotatebox{90}{Exp. Moving Avg.} 
    
    & \rotatebox{90}{pre-trained $224^2$}
    & \rotatebox{90}{fine-tuned $384^2$}  \\
    
    \toprule
    none: \oursbase  & adamw & adamw & \cmark & \xmark &  \cmark & \cmark & \cmark & \cmark & \cmark & \xmark & \xmark & 81.8\stdminus{0.2} & 83.1\stdminus{0.1} \\
    \midrule

    \multirow{2}{*}{\rotatebox{0}{optimizer}}
                      & SGD & \adamwg & \cmarkg & \xmarkg &  \cmarkg & \cmarkg & \cmarkg & \cmarkg & \cmarkg & \xmarkg & \xmarkg & 74.5   &  77.3\\ 
                      & \adamwg & SGD & \cmarkg & \xmarkg &  \cmarkg & \cmarkg & \cmarkg & \cmarkg & \cmarkg & \xmarkg & \xmarkg & 81.8  & 83.1 \\ 
    \midrule
    
    \multirow{6}{*}{\rotatebox{0}{\makebox{\begin{minipage}{2cm}\centering data\\ augmentation \\ ~\end{minipage}}}}
                      & \adamwg & \adamwg & \xmark  & \xmarkg & \cmarkg & \cmarkg & \cmarkg & \cmarkg & \cmarkg & \xmarkg  & \xmarkg & 79.6  & 80.4 \\ 
                      & \adamwg & \adamwg & \xmark  & \cmark  & \cmarkg & \cmarkg & \cmarkg & \cmarkg & \cmarkg & \xmarkg  & \xmarkg & 81.2  & 81.9 \\ 
                      & \adamwg & \adamwg & \cmarkg & \xmarkg & \xmark  & \cmarkg & \cmarkg & \cmarkg & \cmarkg & \xmarkg  & \xmarkg & 78.7  & 79.8 \\ 
                      & \adamwg & \adamwg & \cmarkg & \xmarkg & \cmarkg & \xmark  & \cmarkg & \cmarkg & \cmarkg & \xmarkg  & \xmarkg & 80.0  & 80.6 \\ 
                      & \adamwg & \adamwg & \cmarkg & \xmarkg & \xmark  & \xmark  & \cmarkg & \cmarkg & \cmarkg & \xmarkg  & \xmarkg & 75.8  & 76.7 \\ 
    \midrule
    \multirow{5}{*}{\rotatebox{0}{regularization}}
                      & \adamwg & \adamwg & \cmarkg & \xmarkg & \cmarkg & \cmarkg & \xmark & \cmarkg & \cmarkg & \xmarkg & \xmarkg & \pzo4.3*  &  \pzo0.1\\ 
                      & \adamwg & \adamwg & \cmarkg & \xmarkg & \cmarkg & \cmarkg & \cmarkg & \xmark & \cmarkg &  \xmarkg & \xmarkg & \pzo3.4*  & \pzo0.1 \\ 
                      & \adamwg & \adamwg & \cmarkg & \xmarkg & \cmarkg & \cmarkg & \cmarkg & \cmarkg & \xmark &  \xmarkg & \xmarkg & 76.5   & 77.4 \\ 
                      & \adamwg & \adamwg & \cmarkg & \xmarkg & \cmarkg & \cmarkg & \cmarkg & \cmarkg & \cmarkg & \cmark & \xmarkg & 81.3  & 83.1\\ 
                      & \adamwg & \adamwg & \cmarkg & \xmarkg & \cmarkg & \cmarkg & \cmarkg & \cmarkg & \cmarkg & \xmarkg & \cmark & 81.9 & 83.1\\     \bottomrule
    \end{tabular}}
    \caption{Ablation study on training methods on ImageNet~\cite{Russakovsky2015ImageNet12}. The top row ("none") corresponds to our default configuration employed for \ours. The symbols \cmark and \xmark indicates that we use and do not use the corresponding method, respectively.  
    We report the accuracy scores (\%) after the initial training at resolution 224$\times$224, and after fine-tuning %
    at resolution 384$\times$384. 
    The hyper-parameters are fixed according to Table~\ref{tab:comp_hyperparameters}, and may be suboptimal.\newline 
    {\footnotesize * indicates that the model did not train well, possibly because hyper-parameters are not adapted.} 
    }
    \label{tab:ablation}
\end{table*}

In this section we discuss the \ours training strategy to learn vision transformers in a data-efficient manner. 
We build upon PyTorch~\cite{paszke2019pytorch} and the timm library~\cite{pytorchmodels}\footnote{The timm implementation already included a training procedure that improved the accuracy of ViT-B from 77.91\% to 79.35\% top-1, and trained  on Imagenet-1k with a 8xV100 GPU machine.}.
We provide hyper-parameters as well as an ablation study in which we analyze the impact of each choice. 

\paragraph{Initialization and hyper-parameters.}

Transformers are relatively sensitive to initialization. After testing several options in preliminary experiments, some of them not converging, we follow the recommendation of Hanin and Rolnick~\cite{hanin2018start} to initialize the weights with a truncated normal distribution. 

Table~\ref{tab:comp_hyperparameters} indicates the hyper-parameters that we use by default at training time for all our experiments, unless stated otherwise. 
For distillation we follow the recommendations from Cho et al.~\cite{Cho2019OnTE} to select the parameters $\tau$ and $\lambda$. 
We take the typical values $\tau=3.0$ and $\lambda=0.1$ for the usual (soft) distillation.

\begin{table}[t]
\centering
\scalebox{0.9}
{
\begin{tabular}{lccc}
\toprule
Methods & ViT-B~\cite{dosovitskiy2020image}  & DeiT-B \\
\midrule
Epochs   & 300 & 300     \\
\midrule
Batch size & 4096 & 1024\\
Optimizer & AdamW & AdamW\\
     learning rate       & 0.003 &   $0.0005\times \frac{\textrm{batchsize}}{512} $  \\
     Learning rate decay & cosine & cosine  \\
     Weight decay        & 0.3    & 0.05    \\
     Warmup epochs  & 3.4 & 5       \\
\midrule
     Label smoothing $\varepsilon$ & \xmark & 0.1     \\
     Dropout      & 0.1 & \xmark     \\
     Stoch. Depth & \xmark & 0.1 \\
     Repeated Aug & \xmark & \cmark \\
     Gradient Clip. & \cmark & \xmark \\
\midrule
     Rand Augment  & \xmark        & 9/0.5 \\
     Mixup prob.  & \xmark & 0.8     \\
     Cutmix prob.   & \xmark & 1.0    \\
     Erasing prob.    & \xmark & 0.25    \\
 \bottomrule
\end{tabular}}
\vspace{-8pt}
\caption{
 Ingredients and hyper-parameters for our method and Vit-B.
\label{tab:comp_hyperparameters}}
\end{table}

\paragraph{Data-Augmentation.}

Compared to models that integrate more priors (such as convolutions), transformers require a larger amount of data.
Thus, in order to train with datasets of the same size, we rely on extensive data augmentation.
We evaluate different types of strong data augmentation, with the objective to reach a data-efficient training regime.  

Auto-Augment~\cite{Ekin2018AutoAugment}, Rand-Augment
~\cite{Cubuk2019RandAugmentPA}, and random erasing~\cite{Zhong2020RandomED} improve the results. For the two latter we use the timm~\cite{pytorchmodels} customizations, and after ablation we choose Rand-Augment instead of AutoAugment. 
Overall our experiments confirm that transformers require a strong data augmentation: almost all the data-augmentation methods that we evaluate prove to be useful. One exception is dropout, which we exclude from our training procedure.

\paragraph{Regularization \& Optimizers.}

We have considered different optimizers and cross-validated different learning rates and weight decays. 
Transformers are sensitive to the setting of optimization hyper-parameters. %
Therefore, during cross-validation, we tried 3 different learning rates ($5.10^{-4}, 3.10^{-4}, 5.10^{-5}$) and 3 weight decay ($0.03$, $0.04$, $0.05$).
We scale the learning rate according to the batch size with the formula:
$\mathrm{lr}_{\mathrm{scaled}}=\frac{\mathrm{lr}}{512} \times \mathrm{batch size}$, similarly to Goyal et al.~\cite{Goyal2017AccurateLM} except that we use 512 instead of 256 as the base value.

The best results use the AdamW optimizer with the same learning rates as ViT~\cite{dosovitskiy2020image} but with a much smaller weight decay, as the weight decay reported in the paper hurts the convergence in our setting.

We have employed stochastic depth~\cite{Huang2016DeepNW}, which facilitates the convergence of transformers, especially deep ones~\cite{fan2019reducing,fan2020training}. For vision transformers, they were first adopted in the training procedure by Wightman~\cite{pytorchmodels}. 
Regularization like Mixup~\cite{Zhang2017Mixup} and Cutmix~\cite{Yun2019CutMix} improve performance. We also use repeated augmentation~\cite{berman2019multigrain,hoffer2020augment}, which provides a significant boost in performance and is one of the key ingredients of our proposed training procedure. 

\paragraph{Exponential Moving Average (EMA).} 
We evaluate the EMA of our network obtained after training. 
There are small gains, which vanish after fine-tuning: the EMA model has an edge of is 0.1 accuracy points, but when fine-tuned the two models reach the same (improved) performance.

\paragraph{Fine-tuning at different resolution.}

We adopt the fine-tuning procedure from Touvron et al.~\cite{Touvron2020FixingTT}: 
our schedule, regularization and optimization procedure are identical to that of FixEfficientNet but we keep the training-time data augmentation (contrary to the dampened data augmentation of Touvron et al.~\cite{Touvron2020FixingTT}).
We also interpolate the positional embeddings:  
In principle any classical image scaling technique, like bilinear interpolation, could be used. 
However, a bilinear interpolation of a vector from its neighbors reduces its $\ell_2$-norm compared to its neighbors.
These low-norm vectors are not adapted to the pre-trained transformers and we observe a significant drop in accuracy if we employ use directly without any form of fine-tuning.
Therefore we adopt a bicubic interpolation that approximately preserves the norm of the vectors, before fine-tuning the network with either AdamW~\cite{Loshchilov2017AdamW} or SGD. These optimizers have a similar performance for the fine-tuning stage, see Table~\ref{tab:ablation}. 
By default and similar to ViT~\cite{dosovitskiy2020image} we train \ours models with at resolution $224$ and we fine-tune at resolution $384$.
We detail how to do this interpolation in Section~\ref{sec:position_encoding}. 
However, in order to measure the influence of the resolution we have finetuned \ours at different resolutions. 
We report these results in Table~\ref{tab:res_finetune}.

 \begin{table}[t]
    \centering
    \scalebox{0.955}
    {\small 
    \begin{tabular}{c@{\ }c|c@{\ }|c@{\ }|c@{\ }}
    \toprule
             image       & throughput & \multicolumn{1}{c}{Imagenet~\cite{Russakovsky2015ImageNet12}} & \multicolumn{1}{|c|}{Real~\cite{Beyer2020ImageNetReal}}& \multicolumn{1}{c}{V2~\cite{Recht2019ImageNetv2}}\\ 
     size & (image/s) &  acc. top-1 &  acc. top-1 &  acc. top-1\\
    \toprule

     $160^{2}$ & 609.31   & 79.9 &   84.8   & 67.6 \\
     $224^{2}$ & 291.05    & 81.8 & 86.7 & 71.5 \\
     $320^{2}$ & 134.13 & 82.7 &  87.2 & 71.9 \\
     $384^{2}$ & \pzo85.87 & 83.1 & 87.7 & 72.4 \\

    \bottomrule
    \end{tabular}}
    \caption{Performance of \ours trained at size $224^{2}$ for varying finetuning sizes on ImageNet-1k, ImageNet-Real and ImageNet-v2 matched frequency. 
    \label{tab:res_finetune}}
\end{table}

\paragraph{Training time. }

A typical training of 300 epochs takes 37 hours with 2 nodes or 53 hours on a single node for the \ours-B.%
As a comparison point, a similar training with a RegNetY-16GF~\cite{Radosavovic2020RegNet} (84M parameters) is 20\% slower.
\ours-S and \ourstiny are trained in less than 3 days on 4 GPU.
Then, optionally we fine-tune the model at a larger resolution. This takes 20 hours on a single node (8 GPU) to produce a \oursfix model at resolution 384$\times$384, which corresponds to 25 epochs.
Not having to rely on batch-norm allows one to reduce the batch size without impacting performance, which makes it easier to train larger models.
Note that, since we use repeated augmentation~\cite{berman2019multigrain,hoffer2020augment} with 3 repetitions, we only see one third of the images during a single epoch\footnote{Formally it means that we have 100 epochs, but each is 3x longer because of the repeated augmentations. We prefer to refer to this as 300 epochs in order to have a direct comparison on the effective training time with and without repeated augmentation.}.

\section{Conclusion}
\label{sec:conclusion}

In this paper, we have introduced \ours, which are image transformers that do not require very large amount of data to be trained, thanks to improved training and in particular a novel distillation procedure. 
Convolutional neural networks have optimized, both in terms of architecture  and  optimization during almost a decade, including through extensive architecture search that is prone to overfiting, as it is the case for instance for EfficientNets~\cite{Touvron2020FixingTT}. For \ours we have started the existing data augmentation and regularization strategies pre-existing for convnets, not introducing any significant architectural beyond our novel distillation token. Therefore it is likely that research on data-augmentation more adapted or learned for transformers will bring further gains.  
 
Therefore, considering our results, where image transformers are on par with convnets already, we believe that they will rapidly become a method of choice considering their lower memory footprint for a given accuracy.

We provide an open-source implementation of our method. It is available at
\url{https://github.com/facebookresearch/deit}.
\subsection*{Acknowledgements}

Many thanks to Ross Wightman for sharing his ViT code and bootstrapping training method with the community, as well as for valuable feedback that helped us to fix different aspects of this paper. Thanks to Vinicius Reis, Mannat Singh, Ari Morcos, Mark Tygert, Gabriel Synnaeve, and  other colleagues at Facebook for brainstorming and some exploration on this axis. Thanks to Ross Girshick and Piotr Dollar for constructive comments.

\clearpage

{\small
\bibliographystyle{ieee_fullname}
\bibliography{egbib}

\begin{thebibliography}{10}\itemsep=-1pt

\bibitem{abnar2020transferring}
Samira Abnar, Mostafa Dehghani, and Willem Zuidema.
\newblock Transferring inductive biases through knowledge distillation.
\newblock {\em arXiv preprint arXiv:2006.00555}, 2020.

\bibitem{Hu2017SENet}
Jie~Hu andLi Shen and Gang Sun.
\newblock Squeeze-and-excitation networks.
\newblock {\em arXiv preprint arXiv:1709.01507}, 2017.

\bibitem{ba2016layer}
Jimmy~Lei Ba, Jamie~Ryan Kiros, and Geoffrey~E Hinton.
\newblock Layer normalization.
\newblock {\em arXiv preprint arXiv:1607.06450}, 2016.

\bibitem{berman2019multigrain}
Maxim Berman, Herv{\'{e}} J{\'{e}}gou, Andrea Vedaldi, Iasonas Kokkinos, and
  Matthijs Douze.
\newblock Multigrain: a unified image embedding for classes and instances.
\newblock {\em arXiv preprint arXiv:1902.05509}, 2019.

\bibitem{Beyer2020ImageNetReal}
Lucas Beyer, Olivier~J. H{\'e}naff, Alexander Kolesnikov, Xiaohua Zhai, and
  Aaron van~den Oord.
\newblock Are we done with imagenet?
\newblock {\em arXiv preprint arXiv:2006.07159}, 2020.

\bibitem{carion2020end}
Nicolas Carion, Francisco Massa, Gabriel Synnaeve, Nicolas Usunier, Alexander
  Kirillov, and Sergey Zagoruyko.
\newblock End-to-end object detection with transformers.
\newblock In {\em European Conference on Computer Vision}, 2020.

\bibitem{chen2020generative}
Mark Chen, Alec Radford, Rewon Child, Jeffrey Wu, Heewoo Jun, David Luan, and
  Ilya Sutskever.
\newblock Generative pretraining from pixels.
\newblock In {\em International Conference on Machine Learning}, 2020.

\bibitem{Chen2020UNITERUI}
Yen-Chun Chen, Linjie Li, Licheng Yu, A.~E. Kholy, Faisal Ahmed, Zhe Gan, Y.
  Cheng, and Jing jing Liu.
\newblock Uniter: Universal image-text representation learning.
\newblock In {\em European Conference on Computer Vision}, 2020.

\bibitem{Cho2019OnTE}
J.~H. Cho and B. Hariharan.
\newblock On the efficacy of knowledge distillation.
\newblock {\em International Conference on Computer Vision}, 2019.

\bibitem{Chu2020FeatureSA}
P. Chu, Xiao Bian, Shaopeng Liu, and Haibin Ling.
\newblock Feature space augmentation for long-tailed data.
\newblock {\em arXiv preprint arXiv:2008.03673}, 2020.

\bibitem{Ekin2018AutoAugment}
Ekin~Dogus Cubuk, Barret Zoph, Dandelion Man{\'{e}}, Vijay Vasudevan, and
  Quoc~V. Le.
\newblock Autoaugment: Learning augmentation policies from data.
\newblock {\em arXiv preprint arXiv:1805.09501}, 2018.

\bibitem{Cubuk2019RandAugmentPA}
Ekin~D. Cubuk, Barret Zoph, Jonathon Shlens, and Quoc~V. Le.
\newblock Randaugment: Practical automated data augmentation with a reduced
  search space.
\newblock {\em arXiv preprint arXiv:1909.13719}, 2019.

\bibitem{deng2009imagenet}
Jia Deng, Wei Dong, Richard Socher, Li-Jia Li, Kai Li, and Li Fei-Fei.
\newblock Imagenet: A large-scale hierarchical image database.
\newblock In {\em Conference on Computer Vision and Pattern Recognition}, pages
  248--255, 2009.

\bibitem{devlin2018bert}
Jacob Devlin, Ming-Wei Chang, Kenton Lee, and Kristina Toutanova.
\newblock Bert: Pre-training of deep bidirectional transformers for language
  understanding.
\newblock {\em arXiv preprint arXiv:1810.04805}, 2018.

\bibitem{dosovitskiy2020image}
Alexey Dosovitskiy, Lucas Beyer, Alexander Kolesnikov, Dirk Weissenborn,
  Xiaohua Zhai, Thomas Unterthiner, Mostafa Dehghani, Matthias Minderer, Georg
  Heigold, Sylvain Gelly, et~al.
\newblock An image is worth 16x16 words: Transformers for image recognition at
  scale.
\newblock {\em arXiv preprint arXiv:2010.11929}, 2020.

\bibitem{fan2019reducing}
Angela Fan, Edouard Grave, and Armand Joulin.
\newblock Reducing transformer depth on demand with structured dropout.
\newblock {\em arXiv preprint arXiv:1909.11556}, 2019.
\newblock ICLR 2020.

\bibitem{fan2020training}
Angela Fan, Pierre Stock, Benjamin Graham, Edouard Grave, R{\'e}mi Gribonval,
  Herv{\'e} J{\'e}gou, and Armand Joulin.
\newblock Training with quantization noise for extreme model compression.
\newblock {\em arXiv preprint arXiv:2004.07320}, 2020.

\bibitem{ghering2017convseq2seq}
Jonas Gehring, Michael Auli, David Grangier, Denis Yarats, and Yann~N. Dauphin.
\newblock Convolutional sequence to sequence learning.
\newblock {\em arXiv preprint arXiv:1705.03122}, 2017.

\bibitem{Goyal2017AccurateLM}
Priya Goyal, Piotr Doll{\'a}r, Ross~B. Girshick, Pieter Noordhuis, Lukasz
  Wesolowski, Aapo Kyrola, Andrew Tulloch, Yangqing Jia, and Kaiming He.
\newblock Accurate, large minibatch sgd: Training imagenet in 1 hour.
\newblock {\em arXiv preprint arXiv:1706.02677}, 2017.

\bibitem{hanin2018start}
Boris Hanin and David Rolnick.
\newblock How to start training: The effect of initialization and architecture.
\newblock {\em NIPS}, 31, 2018.

\bibitem{He2016ResNet}
Kaiming He, Xiangyu Zhang, Shaoqing Ren, and Jian Sun.
\newblock Deep residual learning for image recognition.
\newblock In {\em Conference on Computer Vision and Pattern Recognition}, June
  2016.

\bibitem{he2019bag}
Tong He, Zhi Zhang, Hang Zhang, Zhongyue Zhang, Junyuan Xie, and Mu Li.
\newblock Bag of tricks for image classification with convolutional neural
  networks.
\newblock In {\em Conference on Computer Vision and Pattern Recognition}, 2019.

\bibitem{Hendrycks2016GaussianEL}
Dan Hendrycks and Kevin Gimpel.
\newblock Gaussian error linear units (gelus).
\newblock {\em arXiv preprint arXiv:1606.08415}, 2016.

\bibitem{Hinton2015DistillingTK}
Geoffrey~E. Hinton, Oriol Vinyals, and J. Dean.
\newblock Distilling the knowledge in a neural network.
\newblock {\em arXiv preprint arXiv:1503.02531}, 2015.

\bibitem{hoffer2020augment}
Elad Hoffer, Tal Ben-Nun, Itay Hubara, Niv Giladi, Torsten Hoefler, and Daniel
  Soudry.
\newblock Augment your batch: Improving generalization through instance
  repetition.
\newblock In {\em Conference on Computer Vision and Pattern Recognition}, 2020.

\bibitem{Horn2018INaturalist}
Grant~Van Horn, Oisin {Mac Aodha}, Yang Song, Alexander Shepard, Hartwig Adam,
  Pietro Perona, and Serge~J. Belongie.
\newblock The inaturalist challenge 2018 dataset.
\newblock {\em arXiv preprint arXiv:1707.06642}, 2018.

\bibitem{Horn2019INaturalist}
Grant~Van Horn, Oisin {Mac Aodha}, Yang Song, Alexander Shepard, Hartwig Adam,
  Pietro Perona, and Serge~J. Belongie.
\newblock The inaturalist challenge 2019 dataset.
\newblock {\em arXiv preprint arXiv:1707.06642}, 2019.

\bibitem{Hu2018RelationNF}
H. Hu, Jiayuan Gu, Zheng Zhang, Jifeng Dai, and Y. Wei.
\newblock Relation networks for object detection.
\newblock {\em Conference on Computer Vision and Pattern Recognition}, 2018.

\bibitem{Huang2016DeepNW}
Gao Huang, Yu Sun, Zhuang Liu, Daniel Sedra, and Kilian~Q. Weinberger.
\newblock Deep networks with stochastic depth.
\newblock In {\em European Conference on Computer Vision}, 2016.

\bibitem{Cars2013}
Jonathan Krause, Michael Stark, Jia Deng, and Li Fei-Fei.
\newblock 3d object representations for fine-grained categorization.
\newblock In {\em 4th International IEEE Workshop on 3D Representation and
  Recognition (3dRR-13)}, 2013.

\bibitem{Krizhevsky2009LearningML}
Alex Krizhevsky.
\newblock Learning multiple layers of features from tiny images.
\newblock Technical report, CIFAR, 2009.

\bibitem{Krizhevsky2012AlexNet}
Alex Krizhevsky, Ilya Sutskever, and Geoffrey~E. Hinton.
\newblock Imagenet classification with deep convolutional neural networks.
\newblock In {\em NIPS}, 2012.

\bibitem{li2019visualbert}
Liunian~Harold Li, Mark Yatskar, Da Yin, Cho-Jui Hsieh, and Kai-Wei Chang.
\newblock {VisualBERT:} a simple and performant baseline for vision and
  language.
\newblock {\em arXiv preprint arXiv:1908.03557}, 2019.

\bibitem{Li2019SelectiveKN}
Xiang Li, Wenhai Wang, Xiaolin Hu, and Jian Yang.
\newblock Selective kernel networks.
\newblock {\em Conference on Computer Vision and Pattern Recognition}, 2019.

\bibitem{Locatello2020ObjectCentricLW}
Francesco Locatello, Dirk Weissenborn, Thomas Unterthiner, Aravindh Mahendran,
  Georg Heigold, Jakob Uszkoreit, A. Dosovitskiy, and Thomas Kipf.
\newblock Object-centric learning with slot attention.
\newblock {\em arXiv preprint arXiv:2006.15055}, 2020.

\bibitem{Loshchilov2017AdamW}
I. Loshchilov and F. Hutter.
\newblock Fixing weight decay regularization in adam.
\newblock {\em arXiv preprint arXiv:1711.05101}, 2017.

\bibitem{Lu2019ViLBERTPT}
Jiasen Lu, Dhruv Batra, D. Parikh, and Stefan Lee.
\newblock Vilbert: Pretraining task-agnostic visiolinguistic representations
  for vision-and-language tasks.
\newblock In {\em NIPS}, 2019.

\bibitem{Nilsback08}
M-E. Nilsback and A. Zisserman.
\newblock Automated flower classification over a large number of classes.
\newblock In {\em Proceedings of the Indian Conference on Computer Vision,
  Graphics and Image Processing}, 2008.

\bibitem{paszke2019pytorch}
Adam Paszke, Sam Gross, Francisco Massa, Adam Lerer, James Bradbury, Gregory
  Chanan, Trevor Killeen, Zeming Lin, Natalia Gimelshein, Luca Antiga, et~al.
\newblock Pytorch: An imperative style, high-performance deep learning library.
\newblock In {\em Advances in neural information processing systems}, pages
  8026--8037, 2019.

\bibitem{Radosavovic2020RegNet}
Ilija Radosavovic, Raj~Prateek Kosaraju, Ross~B. Girshick, Kaiming He, and
  Piotr Doll{\'a}r.
\newblock Designing network design spaces.
\newblock {\em Conference on Computer Vision and Pattern Recognition}, 2020.

\bibitem{Recht2019ImageNetv2}
B. Recht, Rebecca Roelofs, L. Schmidt, and V. Shankar.
\newblock Do imagenet classifiers generalize to imagenet?
\newblock {\em arXiv preprint arXiv:1902.10811}, 2019.

\bibitem{Russakovsky2015ImageNet12}
Olga Russakovsky, Jia Deng, Hao Su, Jonathan Krause, Sanjeev Satheesh, Sean Ma,
  Zhiheng Huang, Andrej Karpathy, Aditya Khosla, Michael Bernstein,
  Alexander~C. Berg, and Li Fei-Fei.
\newblock Imagenet large scale visual recognition challenge.
\newblock {\em International journal of Computer Vision}, 2015.

\bibitem{shen2020global}
Zhuoran Shen, Irwan Bello, Raviteja Vemulapalli, Xuhui Jia, and Ching-Hui Chen.
\newblock Global self-attention networks for image recognition.
\newblock {\em arXiv preprint arXiv:2010.03019}, 2020.

\bibitem{Simonyan2015VGG}
K. Simonyan and A. Zisserman.
\newblock Very deep convolutional networks for large-scale image recognition.
\newblock In {\em International Conference on Learning Representations}, 2015.

\bibitem{Sun2019VideoBERTAJ}
C. Sun, A. Myers, Carl Vondrick, Kevin Murphy, and C. Schmid.
\newblock Videobert: A joint model for video and language representation
  learning.
\newblock {\em Conference on Computer Vision and Pattern Recognition}, 2019.

\bibitem{sun2017revisiting}
Chen Sun, Abhinav Shrivastava, Saurabh Singh, and Abhinav Gupta.
\newblock Revisiting unreasonable effectiveness of data in deep learning era.
\newblock In {\em Proceedings of the IEEE international conference on computer
  vision}, pages 843--852, 2017.

\bibitem{Szegedy2016RethinkingTI}
Christian Szegedy, V. Vanhoucke, S. Ioffe, Jon Shlens, and Z. Wojna.
\newblock Rethinking the inception architecture for computer vision.
\newblock {\em Conference on Computer Vision and Pattern Recognition}, 2016.

\bibitem{tan2019efficientnet}
Mingxing Tan and Quoc~V. Le.
\newblock Efficientnet: Rethinking model scaling for convolutional neural
  networks.
\newblock {\em arXiv preprint arXiv:1905.11946}, 2019.

\bibitem{Touvron2020GrafitLF}
Hugo Touvron, Alexandre Sablayrolles, M. Douze, M. Cord, and H. J{\'e}gou.
\newblock Grafit: Learning fine-grained image representations with coarse
  labels.
\newblock {\em arXiv preprint arXiv:2011.12982}, 2020.

\bibitem{Touvron2019FixRes}
Hugo Touvron, Andrea Vedaldi, Matthijs Douze, and Herve Jegou.
\newblock Fixing the train-test resolution discrepancy.
\newblock {\em NIPS}, 2019.

\bibitem{Touvron2020FixingTT}
Hugo Touvron, Andrea Vedaldi, Matthijs Douze, and Herv{\'e} J{\'e}gou.
\newblock Fixing the train-test resolution discrepancy: Fixefficientnet.
\newblock {\em arXiv preprint arXiv:2003.08237}, 2020.

\bibitem{Vaswani2017AttentionIA}
Ashish Vaswani, Noam Shazeer, Niki Parmar, Jakob Uszkoreit, Llion Jones,
  Aidan~N. Gomez, Lukasz Kaiser, and Illia Polosukhin.
\newblock Attention is all you need.
\newblock In {\em NIPS}, 2017.

\bibitem{Wang2018NonlocalNN}
X. Wang, Ross~B. Girshick, A. Gupta, and Kaiming He.
\newblock Non-local neural networks.
\newblock {\em Conference on Computer Vision and Pattern Recognition}, 2018.

\bibitem{Wei2020CircumventingOO}
Longhui Wei, An Xiao, Lingxi Xie, Xin Chen, Xiaopeng Zhang, and Qi Tian.
\newblock Circumventing outliers of autoaugment with knowledge distillation.
\newblock {\em European Conference on Computer Vision}, 2020.

\bibitem{pytorchmodels}
Ross Wightman.
\newblock Pytorch image models.
\newblock \url{https://github.com/rwightman/pytorch-image-models}, 2019.

\bibitem{wu2020visual}
Bichen Wu, Chenfeng Xu, Xiaoliang Dai, Alvin Wan, Peizhao Zhang, Masayoshi
  Tomizuka, Kurt Keutzer, and Peter Vajda.
\newblock Visual transformers: Token-based image representation and processing
  for computer vision.
\newblock {\em arXiv preprint arXiv:2006.03677}, 2020.

\bibitem{Xie2019SelftrainingWN}
Qizhe Xie, Eduard~H. Hovy, Minh-Thang Luong, and Quoc~V. Le.
\newblock Self-training with noisy student improves imagenet classification.
\newblock {\em arXiv preprint arXiv:1911.04252}, 2019.

\bibitem{Yuan2019RevisitKD}
L. Yuan, F. Tay, G. Li, T. Wang, and Jiashi Feng.
\newblock Revisit knowledge distillation: a teacher-free framework.
\newblock {\em Conference on Computer Vision and Pattern Recognition}, 2020.

\bibitem{Yun2019CutMix}
Sangdoo Yun, Dongyoon Han, Seong~Joon Oh, Sanghyuk Chun, Junsuk Choe, and
  Youngjoon Yoo.
\newblock Cutmix: Regularization strategy to train strong classifiers with
  localizable features.
\newblock {\em arXiv preprint arXiv:1905.04899}, 2019.

\bibitem{Zhang2017Mixup}
Hongyi Zhang, Moustapha Ciss{\'{e}}, Yann~N. Dauphin, and David Lopez{-}Paz.
\newblock mixup: Beyond empirical risk minimization.
\newblock {\em arXiv preprint arXiv:1710.09412}, 2017.

\bibitem{zhang2020resnest}
Hang Zhang, Chongruo Wu, Zhongyue Zhang, Yi Zhu, Zhi Zhang, Haibin Lin, Yue
  Sun, Tong He, Jonas Muller, R. Manmatha, Mu Li, and Alexander Smola.
\newblock Resnest: Split-attention networks.
\newblock {\em arXiv preprint arXiv:2004.08955}, 2020.

\bibitem{Zhong2020RandomED}
Zhun Zhong, Liang Zheng, Guoliang Kang, Shaozi Li, and Yi Yang.
\newblock Random erasing data augmentation.
\newblock In {\em AAAI}, 2020.

\end{thebibliography}
}

\end{document}